\documentclass{article} 
\pdfoutput=1
\usepackage{nips14submit_e,times}
\usepackage{url}
\usepackage{amsmath}
\usepackage{amssymb}
\usepackage[numbers,square]{natbib}
\usepackage[T1]{fontenc}     
\usepackage{textcomp}

\usepackage{graphicx}
\usepackage{caption}
\usepackage{subcaption}

\newcommand{\eqdef}{\buildrel \text{d{}ef}\over = }
\newcommand{\argmin}{\mathop{\mathrm{argmin}}}
\newcommand{\sgn}{\mathop{\mathrm{sgn}}}
\newcommand{\myldots}{\;\ldots\;}
\newcommand{\vbeta}{\boldsymbol \beta}
\newcommand{\va}{\mathbf a}

\newcommand{\vx}{\mathbf x}
\newcommand{\vy}{\mathbf y}

\newtheorem{algorithm}{Algorithm}
\newtheorem{lemma}{Lemma}
\newcommand{\suchthat}{\;\ifnum\currentgrouptype=16 \middle\fi|\;}

\title{Distributed Coordinate Descent for L1-regularized Logistic Regression}

\author{
Ilya Trofimov \\
Yandex \\
\texttt{trofim@yandex-team.ru} \\
\And
Alexander Genkin \\
AVG Consulting \\
\texttt{alexander.genkin@gmail.com} \\
}

\nipsfinalcopy 

\begin{document}

\maketitle

\begin{abstract}
Solving logistic regression with L1-regularization in distributed settings is an important problem.
This problem arises when training dataset is very large and cannot fit the memory of a single machine.
We present d-GLMNET, a new algorithm solving logistic regression with L1-regularization in the distributed settings.
We empirically show that it is superior over distributed online learning via truncated gradient.
\end{abstract}

\section{Introduction}

Logistic regression with L1-regularization is the method of choice for solving classification and class probability estimation
problems in  text mining, biometrics and clickstream data analysis.
Despite the fact that logistic regression can build only linear separating surfaces, the performance (i.e., testing accuracy) of it,
with proper regularization, has shown to be close to that of nonlinear classifiers such as kernel methods.
At the same time training and testing of linear classifiers is much faster.
It makes the logistic regression a good choice for large-scale problems. A desirable trait of model is sparsity, which is conveniently achieved with L1 or elastic net regularizer.

A broad survey \citep{Yuan2010} suggests that coordinate descent methods are the best choice for L1-regularized logistic regression on the large scale.
Widely used algorithms that fall into this family are: BBR \citep{Genkin2007}, GLMNET \citep{Friedman2010}, newGLMNET \citep{Yuan2012a}.
Software implementations of these methods start with loading the full training dataset into RAM.

Completely different approach is online learning \citep{Balakrishnan2007, Langford2009, Mcmahan2011, Mcmahan2013}.
This kind of algorithms do not require to load training dataset into RAM and can access it sequentially (i.e. reading from disk).
\citet{Balakrishnan2007}, \citet{Langford2009} report that online learning performs well when compared to batch counterparts (BBR and LASSO).

Nowadays we see the growing  number of problems where both the number of examples and the number of features are very large.
Many problems grow beyond the capabilities of a single computer and need to be handled by distributed systems.
Approaches to distributed training of classifiers naturally fall into two groups by the way they split data across computing nodes: by examples \citep{Agarwal2011}
or by features \citep{Peng2013}.
We believe that algorithms that split data by features can achieve better sparsity while retaining similar or better performance and competitive training speed with those that split by examples.
Our experiments so far confirm that belief.

Parallel block-coordinate descent is a natural algorithmic framework if we choose to split by features. The challenge here is how to combine steps from coordinate blocks, or computing nodes,
and how to organize communication. When features are independent, parallel updates can be combined straightforwardly, otherwise they may come into conflict and not yield enough improvement to objective; this has been clearly illustrated by \citet{Bradley2011}.
\citet{Bradley2011} proposed Shotgun algorithm based  on randomized coordinate descent.
They studied how many variables can be updated in parallel to guarantee convergence.
\citet{Ho2013} presented distributed implementation of this algorithm compatible with Stale Synchronous Parallel Parameter Server.

\citet{Richtarik2012} use randomized block-coordinate descent and also exploit partial separability of the objective. The latter relies on sparsity in data, which  is indeed characteristic to many large scale problems. They present theoretical estimates of speed-up factor of parallelization.
\citet{Peng2013} proposed a greedy block-coordinate descent method, which selects the next coordinate to update based on the estimate of the expected
improvement in the objective.
They found their GRock algorithm to be superior over parallel FISTA and ADMM.

In contrast, our approach is to make parallel steps on all blocks, then use combined update as a direction and perform a line search.
We show that sufficient data for line search have the size  $O(n + p)$, where  $n$ is the number of examples, $p$ is the number of features,
so it can be performed on one machine.
Consequently, that's the amount of data sufficient for communication between machines.
Overall, our algorithm fits into the framework of CGD method proposed by \citet{Tseng2007}, which allows us to prove convergence.
Block-coordinate descent on a single machine is performed as a step of GLMNET \citep{Friedman2010}.

When splitting data by examples, online learning comes in handy. A classifier is trained in online fashion on each subset,
then parameters of classifiers are averaged and used as a warmstart for the next iteration, and so on \citep{Agarwal2011, Zinkevich2010}.
We performed an experimental comparison of our algorithm with distributed online learning.


Our main contributions are the following:
\begin{itemize}
\item We propose a new parallel coordinate descent algorithm for  L1-regularized logistic regression and guarantee its convergence (Section 2)
\item We demonstrate how our algorithm can be efficiently implemented on the distributed cluster architecture (Section 3)
\item We empirically show effectiveness of our implementation in comparison with distributed online learning via truncated gradient (Section 4)
\end{itemize}


The C++ implementation of our algorithm, which we call d-GLMNET, is publicly available at \\
 \textit{https://github.com/IlyaTrofimov/dlr}.

\section{Parallel coordinate descent algorithm} 
\label{math}

In case of binary classification the logistic regression estimates the class probability given the feature vector $\vx$
$$
P(y = +1 | \vx) = \frac{1}{1 + \exp(-\vbeta^T \vx)}
$$
This statistical model is fitted by maximizing the log-likelihood (or minimizing the negated log-likelihood) at the training set. Some penalty is often added to avoid overfitting and numerical ill-conditioning.
In our work we consider L1-regularization penalty, which provides sparsity in the model.
Thus fitting the logistic regression with L1-regularization leads to the optimization problem
\begin{equation}
\label{problem}
\vbeta^* = \argmin_{\vbeta \in R^n} f(\vbeta)
\end{equation}
\begin{equation}
\label{target}
f(\vbeta)= L(\vbeta) + \lambda \|\vbeta\|_1
\end{equation}
where $L(\vbeta)$ is the negated log-likelihood
\begin{equation}
\label{target-smooth}
L(\vbeta) = \sum_{i=1}^{n} \log(1 + \exp(-y_i \vbeta^T \vx_i))
\end{equation}
$y_i \in \{-1, +1\}$ are labels, $\vx_i \in \mathbb{R}^p$ are input features, $\vbeta \in \mathbb{R}^p$ is the unknown vector of weights for input features.
We will denote by $nnz$ the number of non-zero entries in all $x_i$.

The first part of the objective - $L(\vbeta)$ is convex and smooth.
The second part is L1-regularization term - $\lambda ||\vbeta||_1$ is convex and separable, but non-smooth.
Hence one cannot use directly efficient optimization techniques like conjugate gradient method or L-BFGS which are often used for logistic regression with L2-regularization.

Our algorithm is based on building local approximations to the objective (\ref{target}).
A smooth part (\ref{target-smooth}) of the objective has quadratic approximation \citep{Friedman2010}
\begin{align}
\label{IRLS}
L_q(\vbeta , \Delta \vbeta) & \eqdef L(\vbeta) + \nabla L(\vbeta)^T \Delta \vbeta + \frac{1}{2} \Delta \vbeta^T \nabla^2 L(\vbeta) \Delta \vbeta \notag \\
& = \frac{1}{2} \sum_{i = 1}^{N} w_i (z_i - \Delta \vbeta^T \vx_i )^2 + C(\vbeta)
\end{align}
where
\begin{align*}
& z_i = \frac{(y_i + 1) / 2 - p(\vx_i)}{p(\vx_i)(1 - p(\vx_i))}
\\
& w_i = p(\vx_i) (1 - p(\vx_i))
\\
& p(\vx_i) = \frac{1}{1 + e^{-\vbeta^T \vx_i}}
\end{align*}
The core idea of GLMNET and newGLMNET is iterative minimization of the penalized quadratic approximation to the objective
\begin{equation}
\label{quad}
\argmin_{\Delta \vbeta} \left\{ L_q(\vbeta, \Delta \vbeta) + \lambda ||\vbeta + \Delta \vbeta||_1 \right\}
\end{equation}
via cyclic coordinate descent. This form (\ref{IRLS}) of approximation allows to make Newton updates of the vector $\vbeta$ without storing the Hessian explicitly.
Also the approximation (\ref{quad}) has a simple closed-form solution with respect to a single variable $\Delta \beta_j$
\begin{equation}
\label{beta-update}
\Delta \beta_j^* = \frac{T \left(\sum_{i=1}^{n} w_i x_{ij} q_i, \lambda \right)}{\sum_{i=1}^{n} w_i x_{ij}^2} - \beta_j
\end{equation}
$$
T(x, a) = \sgn(x) \max(|x| - a, 0)
$$
$$
q_i = z_i - \Delta \vbeta^T \vx_i + (\beta_j + \Delta \beta_j) x_{ij}
$$
In order to adapt the algorithm to the distributed settings we replace the full Hessian with its block-diagonal approximation $\tilde H$.
More formally: let us split $p$ input features into $M$ disjoint sets $S_k$
$$
\bigcup\limits_{k=1}^{M} S_k = \{1, ..., p\}
$$
$$
S_m \cap S_k = \emptyset, k \neq m
$$
Denote by $\tilde{H}$ a block-diagonal matrix
\begin{table}[h]
\begin{equation}
\label{tilde-h}
\centering
{
\begin{tabular}{cll}

\multicolumn{2}{c}{}
$(\tilde H)_{jl} = $ &
$\left\{
\begin{tabular}{l}
$(\nabla^2 L(\vbeta))_{jl}, \; \mbox{if} \; \exists m: \; j, l \in S_m$ \\
0, \; \mbox{otherwise} \\
\end{tabular}
\right.$

\end{tabular}
}
\end{equation}
\end{table}

Let $\Delta \vbeta = \sum_{m=1}^M \Delta \vbeta^m$, where $\Delta \beta^m_j = 0$ if $j \notin S_m$. Then
\begin{align*}
L_q(\vbeta, \Delta \vbeta^m) & = L(\vbeta) + \nabla L(\vbeta)^T \Delta \vbeta^m + \frac{1}{2} \Delta (\vbeta^m)^T \nabla^2 L(\vbeta) \Delta \vbeta^m \\
& = L(\vbeta) + \nabla L(\vbeta)^T \Delta \vbeta^m + \frac{1}{2} \sum_{j,k \in S_m} (\nabla^2 L(\vbeta))_{jk} \Delta \beta_{j} \Delta \beta_{k} \\
& = \frac{1}{2} \sum_{i = 1}^{N} w_i (z_i - (\Delta \vbeta^m)^T \vx_i)^2 + C(\vbeta)
\end{align*}
by summing this equation over $m$
\begin{align}
\label{eq}
\sum_{m=1}^M L_q(\vbeta, \Delta \vbeta^m) &=
\sum_{m=1}^M \left( L(\vbeta) + \nabla L(\vbeta)^T \Delta \vbeta^m + \frac{1}{2} \sum_{j,k \in S_m} (\nabla^2 L(\vbeta))_{jk} \Delta \beta_{j} \Delta \beta_{k} \right) \notag \\
& = M L(\vbeta) + \nabla L(\vbeta)^T \Delta \vbeta + \frac{1}{2} \Delta \vbeta^T  \tilde{H} \Delta \vbeta
\end{align}
From the equation (\ref{eq}) and separability of L1 penalty follows that solving the approximation to the objective
$$
\argmin_{\Delta \vbeta} \left\{ L(\vbeta) + \nabla L(\vbeta)^T \Delta \vbeta + \frac{1}{2} \Delta \vbeta^T  \tilde{H} \Delta \vbeta  + \lambda ||\vbeta + \Delta \vbeta||_1 \right\}
$$
is equivalent to solving $M$ independent sub-problems
\begin{equation}
\label{sub-problem}
\argmin_{\Delta \vbeta^m} \left\{ L_q(\vbeta, \Delta \vbeta^m) + \sum_{j \in S_m} |\beta_j + \Delta \beta^m_j| \suchthat \Delta \beta^m_j = 0 \; \text{if} \; j \notin S_m \right\}
\end{equation}
and can be done in parallel over $M$ machines. This is the main idea of the proposed algorithm \mbox{d-GLMNET}.
We describe a high-level structure of \mbox{d-GLMNET} in the Algorithm \ref{d-glmnet-high}.

\begin{table}[h!]
\begin{algorithm}
\label{d-glmnet-high}
Overall procedure of d-GLMNET \\
$\vbeta \gets 0$ \\
Split \{1,\myldots,p\} into $M$ disjoint sets $S_1,\myldots,S_M$. \\
Repeat until convergence:
\begin{enumerate}
\item \quad Do in parallel over $M$ machines
\item \quad \quad Minimize $L_q(\vbeta, \Delta \vbeta^m) + ||\vbeta + \Delta \vbeta^m||_1$ with respect to $\Delta \vbeta^m$
\item \quad $\Delta \vbeta \gets \sum_{m=1}^{M} \Delta \vbeta^m$
\item \quad Find $\alpha \in (0, 1]$ by the line search procedure (Algorithm \ref{d-glmnet-line-search})
\item \quad $\vbeta \gets \vbeta + \alpha \Delta \vbeta$
\end{enumerate}
return $\vbeta$
\end{algorithm}
\end{table}

The downside of using line search is that it can hurt sparsity.
We compute the regularization path (Section \ref{d-glmnet-protocol}) by running Algorithm \ref{d-glmnet-high} with decreasing L1 penalty,
and the algorithm starts with $\vbeta = 0$, so absolute values of $\vbeta$ tend to increase.
However there may be cases when  $\Delta \beta_j = -\beta_j$ for some $j$ on step 2 of Algorithm \ref{d-glmnet-high}, so  $\beta_j$ can to go back to $0$.
In that case, if line search on step 3 selects $\alpha < 1$,  then the opportunity for sparsity is lost.

To retain the sparsity our algorithm takes two precautions.
First, line search is prevented if $\alpha = 1$ guarantees sufficient decrease in the objective value (step 1 of Algorithm \ref{d-glmnet-line-search}).
Second, there is a complication in the convergence criterion.
It starts by checking if relative decrease in the objective is sufficiently small or maximum number of iteration has been reached.
If that turns out true, the algorithm checks if setting $\alpha$ back to $1$ would not be too much of an increase in the objective.
If that is also true, the algorithms updates $\vbeta$ with $\alpha = 1$ and then stops.


Algoritm \ref{d-glmnet-sub} presents our approach for solving sub-problem (\ref{sub-problem}).
d-GLMNET makes one cycle of coordinate descent over input features for approximate solving (\ref{sub-problem}).
Despite the fact that GLMNET and newGLMNET use multiple passes we found that our approach works well in practice.
We also use $\tilde{H} + \nu I$ with small $\nu = 10^{-6}$ instead of $\tilde{H}$ in (\ref{eq}).
The fact that matrix $\tilde{H} + \nu I$ is positive definite is essential for the proof of convergence (see Section \ref{convergence}).

\begin{table}
\begin{algorithm}
\label{d-glmnet-sub}
Solving quadratic sub-problem at machine $m$\\
$\Delta \vbeta^m \gets 0$ \\
Cycle over $j$ in $S_m$:
\begin{enumerate}
\item \quad Minimize $L_q(\vbeta, \Delta \vbeta^m) + ||\vbeta + \Delta \vbeta^m||_1$ with respect to $\Delta \beta^m_j$ using (\ref{beta-update})
\end{enumerate}
return $\Delta \vbeta^m$
\end{algorithm}
\end{table}

Like in other Newton-like algorithms a line search should be done to guarantee convergence.
The Algorithm \ref{d-glmnet-line-search} describes our line search procedure.
We found that selecting $\alpha_{init}$ by minimizing the objective (\ref{target}) (step 2, Algorithm \ref{d-glmnet-line-search}) speeds up the convergence of the Algorithm \ref{d-glmnet-high}. We used $b = 0.5, \sigma = 0.01, \gamma = 0$ for numerical experiments.
\begin{table}[t!]
\begin{algorithm}
\label{d-glmnet-line-search}
Line search procedure
\begin{enumerate}
\item If $\alpha = 1$ yields sufficient relative decrease in the objective, return $\alpha = 1$.
\item Find $\alpha_{init} = \argmin_{\delta < \alpha \leq 1} f(\vbeta + \alpha \Delta \vbeta)$, $\delta > 0$.
\item Armijo rule: let $\alpha$ be the largest element of the sequence $\{\alpha_{init} b^j\}_{j=0,1,...}$ satisfying
$$
f(\vbeta + \alpha \Delta \vbeta) \leq f(\vbeta) + \alpha \sigma D
$$
where $0 < b < 1, 0 < \sigma < 1, 0 \leq \gamma < 1$, and
$$
D = \nabla L(\vbeta)^T \Delta \vbeta + \gamma \Delta \vbeta^T \tilde{H} \Delta \vbeta + \lambda \left( ||\vbeta + \Delta \vbeta||_1 - ||\vbeta||_1 \right)
$$
return $\alpha$
\end{enumerate}
\end{algorithm}
\end{table}


\subsection{Convergence}
\label{convergence}

Algorithm d-GLMNET  falls into the general framework of block-coordinate gradient descent (CGD) proposed by \citet{Tseng2007}.
CGD is about minimization of a sum of a smooth function and separable convex function: in our case, negated log-likelihood and L1 penalty.
At each iteration CGD solves penalized quadratic approximation problem
\begin{align}
\label{cgd-quad}
\argmin_{\Delta \vbeta} \left\{ L(\vbeta) + \nabla L(\vbeta)^T \Delta \vbeta + \frac{1}{2} \Delta \vbeta^T H \Delta \vbeta + \lambda ||\vbeta + \Delta \vbeta||_1 \right\}
\end{align}
where  $H$ is positive definite, iteration specific. For convergence it also requires that for some $\lambda_{max}, \lambda_{min} > 0$ for all iterations
\begin{equation}
\label{assumption}
\lambda_{min} I \preceq H \preceq \lambda_{max} I
\end{equation}
At each iteration updates are done over some subset of features. (That would always be all features in our case, so the rules of subset selection are irrelevant).
After that a line search by the Armijo rule should be conducted.
Then \citet{Tseng2007} prove that $f(\vbeta)$ converges as least Q-linearly and $\vbeta$ converges at least R-linearly.

d-GLMNET inherits the properties of newGLMNET, for which \citet{Yuan2012a} already proved that it belongs to the CGD framework and inferred the convergence results.
That's why we only give the sketch of the proof, outlining the difference.
newGLMNET algorithm  in (\ref{cgd-quad}) for $H$ uses full Hessian $H = \nabla^2 L(\vbeta) + \nu I$,
and \citet{Yuan2012a} proves (\ref{assumption}) for that.
Instead, d-GLMNET uses block-diagonal approximation  $H = \tilde H + \nu I$ , where $\tilde H$ is defined in (\ref{tilde-h}).
That's why CGD iteration (\ref{cgd-quad}) for the full set of features is block separable and can be parallelized.
To prove (\ref{assumption}) for block-diagonal $H$ denote its diagonal blocks by $H^1,...,H^M$
and represent an arbitrary vector $\vx$ as a concatenation of subvectors of corresponding size: $\vx^T = (\vx_1^T,...,\vx_M^T)$.
Then we have
$$
\vx^T H \vx = \sum_{m=1}^{M} \vx_m^T H^m \vx_m
$$
Notice that $H^m = \nabla^2 L(\vbeta^m) + \nu I$, where $\nabla^2 L(\vbeta^m)$ is a Hessian over the subset of features $S_m$.
So for each $H^m$ property (\ref{assumption}) is already proved in \citep{Yuan2012a}.
That means $\lambda_{min} ||\vx_m||^2 \leq \vx_m^T H^m \vx_m \leq \lambda_{max} ||\vx_m||^2$ for $m=1,...,M$, and we obtain the required
$$
\lambda_{min} ||\vx||^2 \leq \vx^T H \vx \leq \lambda_{max} ||\vx||^2
$$

\section{Scalable software implementation}
\label{software-implementation}

Typically most of datasets are stored in "by example" form, so a transformation to "by feature" form is required for d-GLMNET.
For large datasets this operation is hard to do on a single machine.
We use a Map/Reduce cluster \citep{Dean2004} for this purpose.
This transformation typically takes 1-5\% of time relative to the regularization path calculating (Section \ref{d-glmnet-protocol}).
Training dataset partitioning over machines is done by means of a Reduce operation.
We did not implemented d-GLMNET completely in the Map/Reduce programming model since it is ill-suited for iterative machine learning algorithms \citep{Low2010, Agarwal2011}.

In d-GLMNET machine $m$ solves at each iteration the sub-problem (\ref{sub-problem}).
The machine $m$ stores the part $X_m$ of training dataset corresponding to a subset $S_m$ of input features.
$X_m = \{L_j | j \in S_m\}$ where $L_j = \{(i, x_{ij}) | x_{ij} \ne 0\}$.
Our program expects that input file is already in "by feature" representation, see Table \ref{tbl:input-file}.
\begin{table}[h!]
\centering
\caption{Input file format}
\label{tbl:input-file}
\begin{tabular}{|c|c|c|c|c|c|c|}
  \hline
  \small{feature\_id} & \small{(example\_id, value)} & \small{(example\_id, value)} & ... & \small{feature\_id} & \small{(example\_id, value)} & ... \\
  \hline
\end{tabular}
\end{table}
This format of input file allows to read training dataset sequentially from the disk and make coordinate updates (\ref{beta-update}) while solving sub-problem (\ref{sub-problem}).
Our program stores into the RAM only vectors: $\vy$, $(\exp(\vbeta^T x_i))$, $(\Delta \vbeta^T x_i)$, $\vbeta$, $\Delta \vbeta$.
Thus the total memory footprint of our implementation is $O(n + p)$.

Algorithm \ref{program} presents a high-level structure of our software implementation.
We consider this as a general framework for distributed block-coordinate descent, which can be used with various types of updates during step 2.
\begin{algorithm}
\label{program}
Distributed coordinate descent \\
Repeat until convergence:
\begin{enumerate}
\item Do in parallel over $M$ machines
\item \quad Read part of training dataset $X_m$ sequentially; make updates of $\Delta \vbeta^m$, $(\Delta (\vbeta^m)^T x_i))$
\item Sum up vectors $\Delta \vbeta^m$, $(\Delta (\vbeta^m)^T x_i)$ using MPI\_AllReduce: \footnote{We used an implementation from the Vowpal Wabbit project\\  $https://github.com/JohnLangford/vowpal\_wabbit$}
\item \quad $\Delta \vbeta \gets \sum_{m=1}^{M} \Delta \vbeta^m$
\item \quad $(\Delta \vbeta^T \vx_i) \gets \sum_{m=1}^{M} (\Delta (\vbeta^m)^T \vx_i)$
\item Find step size $\alpha$ using line search (Algorithm \ref{d-glmnet-line-search})
\item $\vbeta \gets \vbeta + \alpha \Delta \vbeta$,
\item $(\exp(\vbeta^T x_i)) \gets (\exp(\vbeta^T x_i + \alpha \Delta \vbeta^T x_i))$
\end{enumerate}
\end{algorithm}
Sequential data reading from disk instead of RAM may slow down the program in case of smaller datasets, but it makes the program more scalable.
Also it conforms to the typical pattern of a multi-user cluster system: large disks, many jobs started by different users are running simultaneously.
Each job might process large data but it is allowed to use only a small part of RAM at each machine.

Solving sub-problem (\ref{sub-problem}) during step 2 in Algorithm \ref{program} requires $O(nnz)$ operations and it is well suited for large and sparse datasets. The communication cost during step 3 in Algorithm \ref{program} is $O((n + p) \ln M)$. A logarithmic term arises because machines communicate via a tree structure during MPI\_AllReduce.


\section{Numerical experiments}
\label{numerical}
\subsection{Datasets and experimental settings}

\begin{table}
\centering
\caption{Datasets summary}
\label{tbl:datasets}
\begin{tabular}{|c|c|c|c|c|c|}
  \hline
  dataset & size & \#examples (train/test) & \#features & nnz & avg nonzeros \\
  \hline
  \hline
  epsilon     &  12 Gb & $0.4 \times 10^6$ / $0.1 \times 10^6$   & $2000$               & $8.0 \times 10^8$  & 2000 \\
  \hline
  webspam     &  21 Gb & $0.315 \times 10^6$ / $0.035 \times 10^6$ & $16.6 \times 10^6$ & $1.2 \times 10^9$  & 3727 \\
  \hline
  dna & 71 Gb & $45 \times 10^6$ / $5 \times 10^6$                 & $800$              & $9.0 \times 10^9$  & 200 \\
  \hline
\end{tabular}
\end{table}

We used three datasets for numerical experiments.
These datasets are from the Pascal Large Scale Learning Challenge 2008 \footnote{http://largescale.ml.tu-berlin.de/}

\begin{itemize}
\item \textbf{epsilon} - A synthetic dataset, we used preprocessing and train/test splitting from http://www.csie.ntu.edu.tw/ \texttildelow cjlin/libsvmtools/datasets/binary.html
\item \textbf{webspam} - Webspam classification problem, we used preprocessing and train/test splitting from  http://www.csie.ntu.edu.tw/ \texttildelow cjlin/libsvmtools/datasets/binary.html
\item \textbf{dna} - Splice cite recognition problem. We did the same preprocessing as in challenge (see ftp://largescale.ml.tu-berlin.de/largescale/dna/) and did train/test splitting
\end{itemize}

%
%
The datasets are summarized in Table \ref{tbl:datasets}.
Numerical experiments were carried out at 16 multicore blade servers having Intel(R) Xeon(R) CPU E5-2660 2.20GHz, 32 GB RAM, connected by Gigabit Ethernet.
Each server ran one instance of d-GLMNET or Vowpal Wabbit at once.

\subsection{Experimental protocol for d-GLMNET}
\label{d-glmnet-protocol}

\begin{figure}[t]
        \label{fig:results}
        \centering
        \begin{subfigure}[t]{0.5\textwidth}
                \includegraphics[width=\textwidth]{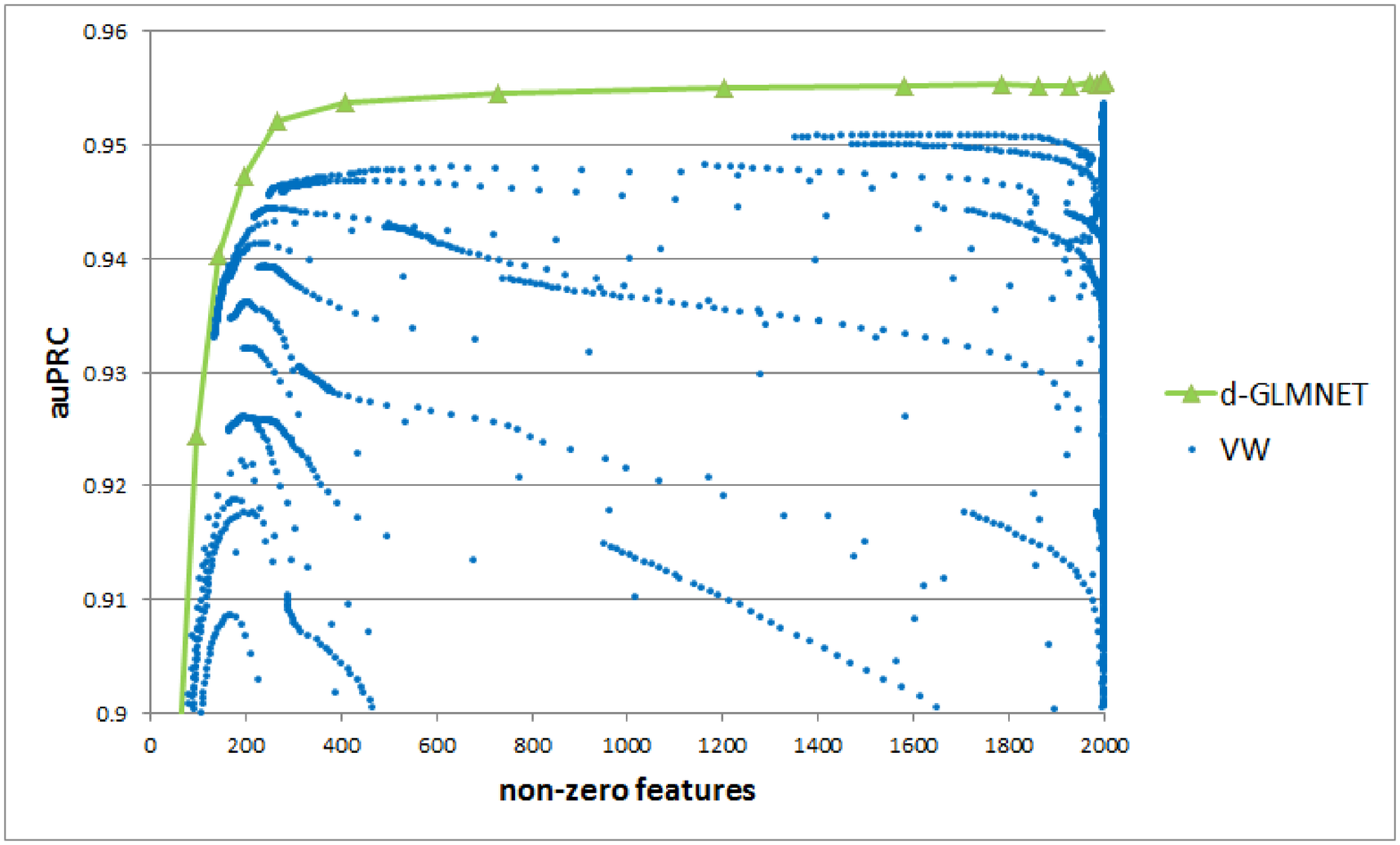}
                \caption{epsilon}
                \label{fig:epsilon}
        \end{subfigure}%
        ~ 
        \begin{subfigure}[t]{0.5\textwidth}
                \includegraphics[width=\textwidth]{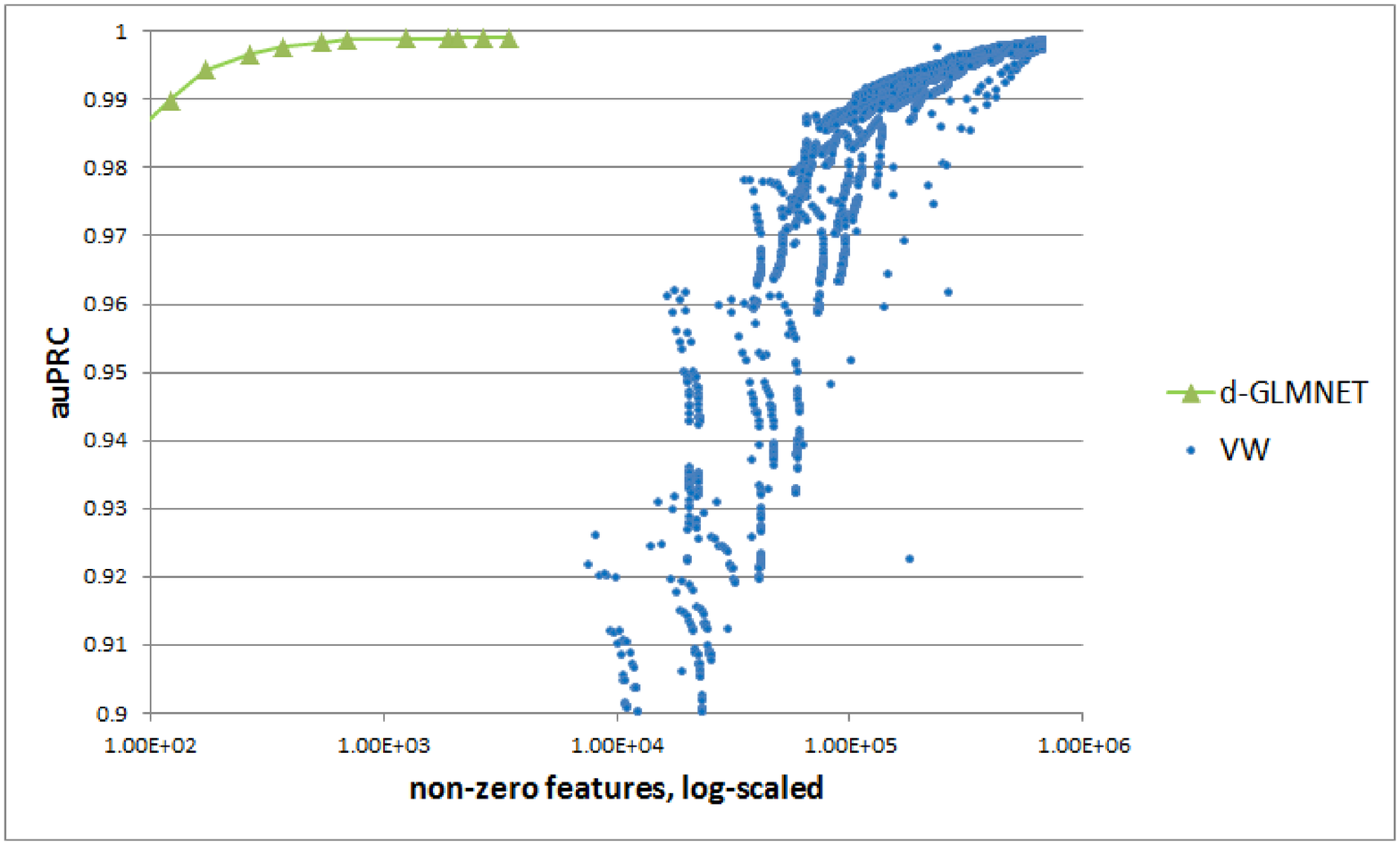}
                \caption{webspam}
                \label{fig:webspam}
        \end{subfigure}

        \begin{subfigure}[t]{0.5\textwidth}
                \includegraphics[width=\textwidth]{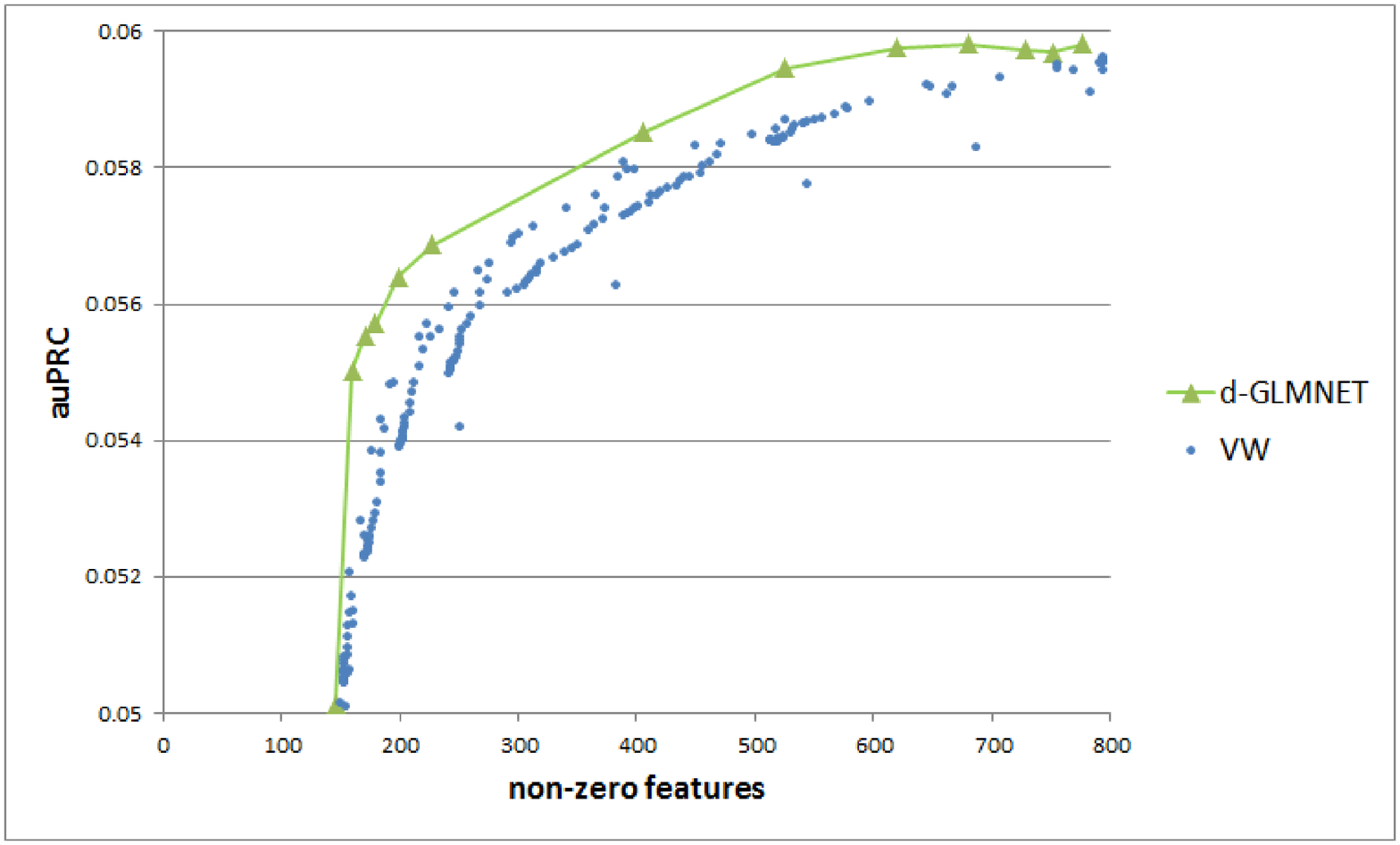}
                \caption{dna}
                \label{fig:dna}
        \end{subfigure}%

        \caption{Testing quality (area under Precision-Recall curve) versus non-zero entries count in $\vbeta$}
\end{figure}

We tested d-GLMNET by solving the problem (\ref{problem}) for a set of regularization parameters, see Algorithm \ref{alg:reg-path}.
\begin{algorithm}
Computing the regularization path
\label{alg:reg-path}
\item Find $\lambda_{max}$ for which entire vector $\vbeta = 0$.
\item For $i = 1$ to $20$
\item \quad Solve (\ref{problem}) with $\lambda = \lambda_{max} * 2 ^ {-i}$ using previous $\vbeta$ as a warmstart
\end{algorithm}
For each $\lambda$ we calculated for a corresponding final $\vbeta$ the testing quality and the number of non-zero entries.
For the "dna" dataset we tested 4 additional regularization parameters $\lambda \in [2730.7, 5461.3]$ because of low density of points in the region with $100-300$ non-zero features (Figure \ref{fig:dna}).

\subsection{Experimental protocol for distributed online learning via truncated gradient}
\label{comparison_with_vw}
We compared d-GLMNET with the distributed variant of online learning via truncated gradient.
The online learning via truncated gradient was presented in \citep{Langford2009}.
An idea for adapting it to the distributed settings was presented in \citep{Agarwal2011}.
We used the first part of \citep[Algorithm 2]{Agarwal2011} which proposes to compute a weighted average of classifiers trained at $M$ machines independently.
The second part of this algorithm takes the result of the first part as a warmstart for L-BFGS.
As we pointed out earlier L-BFGS it not applicable for solving logistic regression with L1-regularization.
This algorithm requires training dataset partitioning by examples over $M$ machines.


The Algorithm 2 from \citep{Agarwal2011} is implemented in the Vowpal Wabbit project \footnote{$https://github.com/JohnLangford/vowpal\_wabbit$, we used version 7.5}.
We tested the same set of regularization parameters as for d-GLMNET, i.e  $\lambda \in \{\lambda_{max} 2^{-1}, \lambda_{max} 2^{-2}, ..., \lambda_{max} 2^{-20}\}$
\footnote{The parameter $\lambda$ in (\ref{target}) is related to the option \textit{- -l1 arg} in Vowpal Wabbit by equation $arg = \lambda / n$ where $n$ is the number of training examples}.
Since online learning has many free parameters we made a full search for "epsilon" and "webspam" datasets.
We tested jointly learning rates (raging from $0.1$ to $0.5$), decays of the learning rate (raging from 0.5 to 0.9) for each $\lambda$ and allowed Vowpal Wabbit to make $50$ passes of online learning.
After each pass we saved a vector $\vbeta$.
After training we evaluated a quality of all classifiers at the test set and counted the number of non-zero entries in $\vbeta$.

For the biggest dataset "dna" we did 25 passes and used default learning rate ($0.1$) and decay ($0.5$).
We also tested additional range of regularization parameter $\lambda \in \{10.7, 10.7 \times 2^{-1}, ..., 10.7 \times 2^{-9}\}$ since Vowpal Wabbit produced only very sparse classifiers with low testing quality.

\subsection{Results}
Figure 1 demonstrates results of the experiments: area under Precision-Recall curve on the test set against the number of non-zero components in the $\vbeta$.
We compare results for the whole regularization path of d-GLMNET and each parameter combination and pass number for Vowpal Wabbit.
The d-GLMNET algorithm is a clear winner: for each data set,  each degree of sparsity, it yields the same or better testing quality.
We notice that for online learning different combinations of  parameters yield very different results.
Online learning is often advertised as a very fast method, but the need to perform a search of good parameters lessens this advantage.
At the same time the d-GLMNET algorithm has no free parameters except a regularization coefficient.

Table 3 presents execution times for the whole regularization pass for each dataset, total number of iterations, and average time per iteration.
We found that linear search does not hurt much the performance - it takes 5-25\% time at different datasets.
There is no direct time comparison between d-GLMNET and Vowpal Wabbit because of the parameter search for the latter.
The last column in the table gives average time per iteration for Vowpal Wabbit: this can be compared to the same number for d-GLMNET,
because one iteration for both algorithms corresponds to one full pass over the training data set, and has the same computational complexity $O(nnz)$.
\begin{table}
\centering
\caption{Execution times}
\label{tbl:timing}
\begin{tabular}{|c|c|c|c|c|c|c|}
  \hline
  \multicolumn{1}{|c|}{} & \multicolumn{4}{|c|}{d-GLMNET} & \multicolumn{1}{|c|}{Vowpal Wabbit}\\
  \hline
  dataset & \#iter & time, sec & linear search & avg time per iter, sec & avg time per iter, sec \\
  \hline
  \hline
  epsilon     & 182 & 1667  & 5\% & 9   & 30 \\
  \hline
  webspam     & 269 & 6318  & 6\% & 23  & 50 \\
  \hline
  dna         & 123 & 17626 & 25\% & 143 & 59 \\
  \hline
\end{tabular}
\end{table}

\subsubsection*{Acknowledgments}
We would like to thank John Langford for the advices on Vowpal Wabbit and Ilya Muchnik for his continuous support.
\bibliographystyle{apalike}
\bibliography{paper}

\end{document}